\documentclass[letterpaper, 10 pt, conference]{ieeeconf}  

\IEEEoverridecommandlockouts                            

\usepackage{graphics} 
\usepackage{epsfig} 
\usepackage{mathptmx} 
\usepackage{times} 
\usepackage{cite}
\usepackage{amsmath,amssymb,amsfonts}
\usepackage{algorithmic}
\usepackage[graphicx]{realboxes}
\usepackage{textcomp}
\usepackage[table,xcdraw]{xcolor}
\usepackage{multirow}
\usepackage{array}
\usepackage{tikz}
\usepackage{siunitx}
\usepackage{lscape}
\usepackage{adjustbox}
\usepackage{tabularx}
\usepackage{makecell}
\usepackage{mleftright}
\usepackage{etoolbox}
\usepackage{xcolor}
\usepackage[normalem]{ulem}
\usepackage{tablefootnote}
\usepackage{verbatim}
\usepackage{graphicx}
\usepackage{gensymb}
\usepackage{titlesec}

\title{\LARGE \bf
Learning Free-Form Deformation for 3D Face Reconstruction from In-The-Wild Images
}

\author{Harim Jung$^{1}$, Myeong-Seok Oh$^{2}$, and Seong-Whan Lee$^{1}$
\thanks{This work was supported by Institute of Information \& communications Technology Planning \& Evaluation (IITP) grant funded by the Korea government (MSIT) (No. 2019-0-00079,  Artificial Intelligence Graduate School Program (Korea University)).}%
\thanks{$^{1}$H. Jung and S.-W. Lee are with the Department of Artificial Intelligence, Korea University, Seoul 02841, South Korea.
{\tt\small hr\_jung@korea.ac.kr and sw.lee@korea.ac.kr}}%
\thanks{$^{2}$M.-S. Oh is with the Department of Computer and Radio Communications Engineering, Korea University, Seoul 02841, South Korea.
{\tt\small ms\_oh@korea.ac.kr}}%
}

\begin{document}

\maketitle
\thispagestyle{empty}
\pagestyle{empty}

\begin{abstract}
The 3D Morphable Model (3DMM), which is a Principal Component Analysis (PCA) based statistical model that represents a 3D face using linear basis functions, has shown promising results for reconstructing 3D faces from single-view in-the-wild images. However, 3DMM has restricted representation power due to the limited number of 3D scans and global linear basis. To address the limitations of 3DMM, we propose a straightforward learning-based method that reconstructs a 3D face mesh through Free-Form Deformation (FFD) for the first time. FFD is a geometric modeling method that embeds a reference mesh within a parallelepiped grid and deforms the mesh by moving the sparse control points of the grid. As FFD is based on mathematically defined basis functions, it has no limitation in representation power. Thus, we can recover accurate 3D face meshes by estimating the appropriate deviation of control points as deformation parameters. 
Although both 3DMM and FFD are parametric models, deformation parameters of FFD are easier to interpret in terms of their effect on the final shape.
This practical advantage of FFD allows the resulting mesh and control points to serve as a good starting point for 3D face modeling, in that ordinary users can fine-tune the mesh by using widely available 3D software tools. Experiments on multiple datasets demonstrate how our method successfully estimates the 3D face geometry and facial expressions from 2D face images, achieving comparable performance to the state-of-the-art methods.

\end{abstract}
\begin{keywords}
3D face reconstruction, Free-form deformation, 3D morphable model
\end{keywords}

\section{INTRODUCTION}

The task of inferring the 3D face geometry and appearance from 2D images has been extensively explored in computer vision and graphics research, as it is essential for many face-related tasks and applications such as face recognition, anti-spoofing, tracking, virtual and augmented reality, animation, and gaming. Recovering a 3D face mesh from a single unconstrained in-the-wild image is especially challenging, due to large head pose variations, extreme expressions, occlusions, lighting conditions, and complex backgrounds.

Research in computer vision has quickly advanced and has been widely used in various applications \cite{yang2007reconstruction, he2016deep, lee1990translation, lim2000text, zhu2016face}. Recently, considerable improvement has been made in 3D face reconstruction, with the help of Convolutional Neural Networks (CNNs).
Previous research on 3D face reconstruction can be mainly categorized into model-based and model-free methods. The statistical PCA-based face model, so-called the 3D Morphable Model (3DMM), has established the foundations of model-based methods. 3DMM is a globally linear model, where the face shape is represented as a linear combination of basis meshes obtained from a set of collected 3D face scans. Lately, many face reconstruction methods began to employ CNNs to regress 3DMM parameters \cite{zhu2016face, guo2020towards, liu2017dense, sanyal2019learning, tewari2017mofa}. 

On the other hand, model-free methods do not rely on a predefined face model but directly regress 3D vertices using volumetric representations \cite{jackson2017large} or UV position maps \cite{feng2018joint} for example. The idea of nonlinear 3DMM \cite{zhou2019dense, tran2018nonlinear} was also introduced, where the nonlinear decoder of a deep neural network maps the shape and texture parameters to the 3D shape and texture. Since the decoder forms the final mesh through direct vertices regression rather than the parameters, it can be considered as model-free. While 3DMM has a model space restricted to the distribution of a specific set of 3D face scans, model-free methods do not have limited representation power. Nevertheless, they tend to be computationally inefficient due to the high degrees of freedom.

In order to address the limitations of model-based and model-free methods, we propose a learning-based 3D face reconstruction method that uses Free-Form Deformation (FFD) \cite{hsu1992direct}, for the first time to the best of our knowledge. FFD is a well-known geometric modeling technique. It embeds a reference mesh within a parallelepiped grid and deforms the mesh by shifting the control points of the grid. Since our FFD-based method does not have limited representation power nor excessively high degrees of freedom, it can be seen to fall into the category between model-based and model-free methods. It is ``model-free" in the sense that it does not reflect actual face scans as in 3DMM. Our method tends to have relatively more parameters to learn than 3DMM-based methods for this reason. On the other hand, it is ``model-based" in the sense that it does not directly use vertices to represent a mesh but lower dimensional control points and mathematically defined basis functions. Thus, it requires less parameters than direct vertices regression. Our goal is to train the network to find the proper deviation of control points, which deforms the reference mesh to be similar to the target face. We summarize our main contributions as follows:

\begin{itemize}
\item 
We explore how FFD can be applied to 3D face meshes in the context of deep learning. Our method attempts to discover the appropriate number of control points, their distribution, and their range of influence over vertices.

\item 
While 3DMM-based methods tend to be restricted in the spanned model space of linear bases, our method has no limit in representation power and generalizes well for unseen faces, as FFD is based on mathematically defined basis functions rather than PCA basis functions.

\item 
Our method can be readily utilized for practical purposes, since the reconstructed mesh and control points can provide a solid starting point for 3D face modeling and can be easily modified for detailed adjustments by using widely available 3D software tools.

\item 
Our method either outperforms or achieves comparable results to existing 3D face reconstruction methods, both in quantitative and qualitative experiments.

\end{itemize}

\section{RELATED WORK}
\subsection{3D Face Reconstruction from a Single Image}
Since Blanz \emph{et al}. \cite{blanz1999morphable} proposed the 3DMM, it has played a dominant role in 3D face modeling and reconstruction. Later on, more advanced variants of 3DMM were designed utilizing larger 3D scan databases and higher dimensional basis \cite{cao2013facewarehouse, paysan20093d, booth20163d, shin2012morphable}. 
Most recent methods use CNNs to regress the 3DMM parameters in a supervised manner, which are used to reconstruct 3D faces \cite{zhu2016face, guo2020towards, liu2017dense}.
Moreover, 3DMM parameters may be found through unsupervised methods \cite{sanyal2019learning, tewari2017mofa} without the help of training data. Sanyal \emph{et al}. \cite{sanyal2019learning} proposed a novel shape consistency loss that induces the face shape to be similar for images of the same person and dissimilar for different people. 

Model-based methods, however, have limited representation power and model-free methods \cite{jackson2017large, feng2018joint, deng2020retinaface} were proposed to overcome this limitation. Jackson \emph{et al}. \cite{jackson2017large} proposed to map image pixels to a volumetric representation, while Feng \emph{et al}. \cite{feng2018joint} developed a UV position map to represent the 3D shape. In a similar fashion, Deng \emph{et al}. \cite{deng2020retinaface} directly regressed 3D vertex coordinates in the image space.
Tran \emph{et al}. \cite{tran2018nonlinear} first introduced the concept of nonlinear 3DMM, where the decoders act as nonlinear models that map the parameters to the actual 3D shape and texture. 

\subsection{3D Shape Reconstruction using Free-Form Deformation}
There have been previous works that attempted to learn FFD for 3D shape reconstruction for rigid objects. Kuryenkov \emph{et al}. \cite{kurenkov2018deformnet} first searched for the nearest shape template from a database and applied FFD to manipulate the template to match the target image. Jack \emph{et al}. \cite{jack2018learning} proposed a similar approach where they learned to deform points sampled from high-quality meshes. 

To the best of our knowledge, FFD has never been applied to learning-based 3D face reconstruction, which is fairly different from general object reconstruction. The object reconstruction methods \cite{kurenkov2018deformnet, jack2018learning} incorporated FFD using Bernstein basis functions, which is appropriate for rigid shapes, as its control points impose global influence on vertices. This method is not necessarily suitable for deformable shapes such as the human face, so we further experiment with FFD using B-spline basis functions.
Moreover, they did not consider the pose of objects by using images obtained from the same viewpoint.
However, considering the head orientation is important in the face domain, since human faces tend to have more obvious changes in pose. For this reason, we train our model to not only regress deformation parameters but also camera projection parameters.

\section{METHOD}

\begin{figure}[t]
    \centering
    \includegraphics[width=0.5\columnwidth]{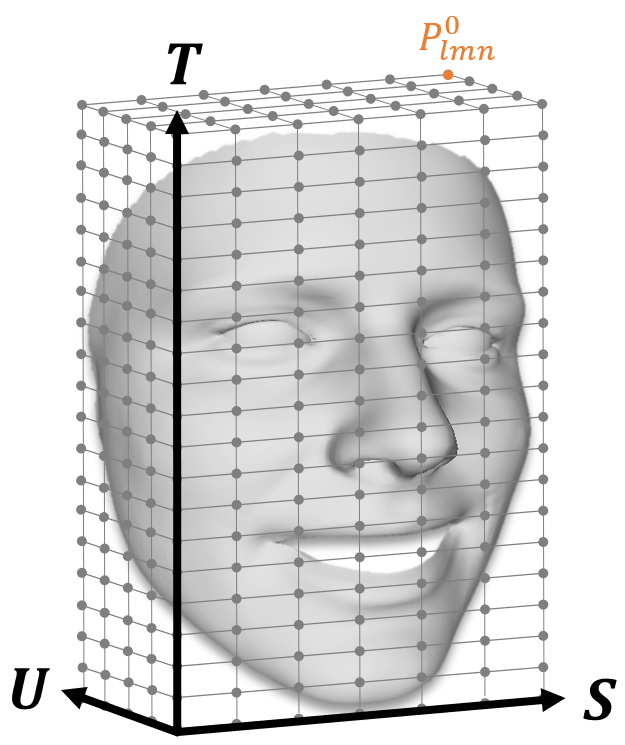}
    \caption{Reference face mesh with frontal pose embedded in the 3D parallelepiped grid of 700 control points ($l=6$, $m=19$, $n=4$). Each point on the intersections represents a control point and $P_{lmn}^0$ refers to the last control point, where $l, m, n$ are the number of divisions along the $S,T,U$ axes respectively. The control points are displaced from their original positions to deform the shape of the embedded reference mesh to match the target face.}\label{fig:1}
\end{figure}

\begin{figure*}[t]
\centering
\includegraphics[width=1.0\linewidth]{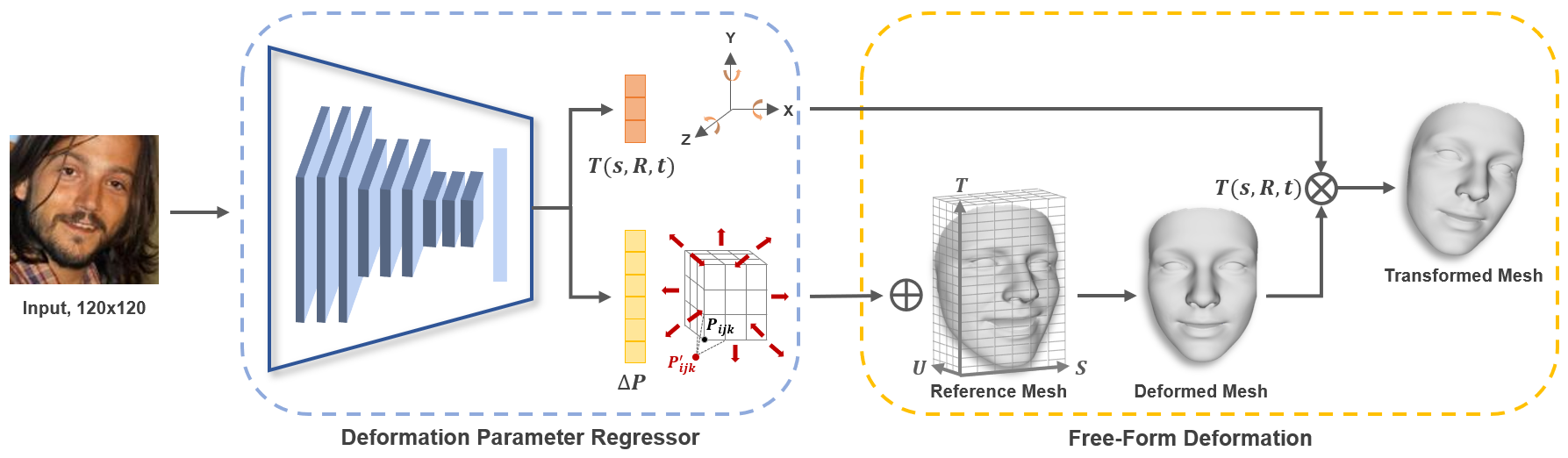}
\caption{Overview of our method. The Deformation Parameter Regressor includes the ResNet-50 backbone, which regresses the pose parameters $T(s,R,t)$ and the deformation parameters $\Delta P$ of 700 control points. The grid on the right of $\Delta P$ illustrates the possible deviation of control points in all directions of the red arrows and as an example, the left-bottom control point $P_{ijk}$ could be moved to the position of $P^{'}_{ijk}$, which would effect the reference mesh accordingly. The Free-Form Deformation part takes in the predicted $\Delta P$, which is added to the original control points $P^0$ of the reference mesh, and the displaced control points are multiplied to $B^0$, to obtain the deformed mesh in the world coordinate system, as in Eq. (\ref{eq:2}). $T(s,R,t)$ is a 3D scaled orthographic projection ($3\times4$ affine transformation), which is multiplied to the deformed mesh to output a transformed mesh in the camera coordinate system. $R, t$ are the rotation and translation parameters and $s$ is the scale factor applied to all $x,y,z$ directions.}\label{fig:2}
\end{figure*}

\subsection{Deforming 3D Face Mesh with Free-Form Deformation}
The goal of our method is to reconstruct the dense 3D face geometry from a single in-the-wild image. Since there are numerous vertices in a mesh, using all vertices as free variables to represent a mesh would be computationally inefficient. The more fundamental problem with this approach, however, is that these variables are not all free because the vertices should be constrained by each other to construct a mesh. Thus, we need to define new variables whose degrees of freedom are just enough to represent a mesh. 
For this purpose, we use FFD to represent and manipulate 3D meshes.
\par

FFD is a shape modification method that has been widely used for geometric modeling in computer graphics and is supported by almost all 3D softwares. It embeds a reference mesh in a parallelepiped grid and deforms it by moving the control points of the grid. Each vertex of a deformed mesh is computed by a linear combination of control points by means of coefficient basis functions \cite{hsu1992direct}. Originally, FFD was implemented by using Bernstein polynomials as basis functions \cite{sederberg1986free}, which we refer to as Bernstein FFD. In this approach, each vertex of the mesh is influenced by all control points of the grid. Therefore, we further experiment with FFD using B-spline basis functions \cite{hsu1992direct}, in which each vertex is influenced by only a small number of neighbor control points. We refer to this method as B-spline FFD.
\par

An arbitrary volume $v(s,t,u)$ can be represented by taking a linear combination of control points using B-spline polynomials as coefficients, defined as follows:

\begin{equation}\label{eq:1}
    \begin{aligned}
        &v(s,t,u) = \sum_{i=0}^{l} \sum_{j=0}^{m} \sum_{k=0}^{n} B_{ijk}(s,t,u) P_{ijk}, \\
        &B_{i,j,k}(s,t,u) =  \sum_{i=0}^{l} \sum_{j=0}^{m} \sum_{k=0}^{n} B_{i,p}(s) B_{j,p}(t)  B_{k,p}(u), \\
        & \textrm{where} \quad 0\leq s,t,u \leq 1.
    \end{aligned}
\end{equation}
Here $B_{i,p}(s)$ is a B-spline basis function of degree $p$ defined over the knot spans dividing the range of $s$. Basis functions $B_{j,p}(t)$ and  $B_{k,p}(u)$ are similarly defined. In this work, we use B-spline functions of degree 3, that is, $p=3$. The number of control points are $(l+1), (m+1)$, and $(n+1)$ in each direction. The control points $P_{ijk}$ are distributed in a lattice structure where the space between the control points is nonuniform in general, but we particularly use a uniform grid with different dimensions on each axis.

We choose one of the mesh data from 300W-LP \cite{zhu2016face} as the reference mesh, as shown in Fig. \ref{fig:1}. Note that the reference mesh can be any face mesh in the world coordinate system, since the goal is to find the appropriate deformation to reach the target mesh. To represent the reference mesh in the context of FFD, we need to define a parametric representation of the mesh by obtaining the values of the parameters $(s_q,t_q,u_q)$ corresponding to each vertex $(x_q,y_q,z_q)$. Considering the relationship between each vertex and the parameters expressed in the following Eq. (\ref{eq:2}), we can obtain these parameters by solving the nonlinear equations:

\begin{equation}\label{eq:2}
    \begin{aligned}
   (x_q,y_q,z_q) &= v(s_q,t_q,u_q) \\
   & = \sum_{i=0}^{l} \sum_{j=0}^{m} \sum_{k=0}^{n} B^0_{ijk}(s_q,t_q,u_q) P^{0}_{ijk},\\
   & \ \ \ \text{for each} \  (x_q,y_q,z_q) \in \emph{RefMesh},
    \end{aligned}
\end{equation}
where \emph{RefMesh} refers to the reference mesh and $P^0$ refers to the control points of the undeformed, initial grid embedding the reference mesh.
$B^{0}_{ijk}(s_q,t_q,u_q)$ is the B-spline coefficient function computed with respect to the reference mesh, which has the property that given a vertex $(x_q,y_q,z_q)$, it has a large value if the control point $P_{ijk}$ is close to the vertex and a small value if it is far from the vertex.

By representing Eq. (\ref{eq:2}) in a matrix multiplication form, the reference mesh $V^{0}$ is represented in terms of the coefficient matrix $B^{0}$ and the control points $P^{0}$ as follows:

\begin{equation}\label{eq:3}
    V^{0} = B^{0}P^{0},
\end{equation}
that is,

\small\begin{equation}\label{eq:4}
    \begin{bmatrix}
        v^{0}(s_1,t_1,u_1)\\
        \vdots \\
        v^{0}(s_N, t_N,u_N)
    \end{bmatrix}
    \hspace{-0.1cm}=\hspace{-0.1cm}
    \begin{bmatrix}
        B^{0}_{000}(s_1,t_1,u_1) \hdots B^{0}_{lmn}(s_1,t_1,u_1) \\
        \vdots \\
        B^{0}_{000}(s_N, t_N,u_N) \hdots B^{0}_{lmn}(s_N, t_N,u_N)
    \end{bmatrix}\hspace{-0.1cm}
    \begin{bmatrix}
        P^{0}_{000}\\
        \vdots \\
        P^{0}_{lmn}
    \end{bmatrix}.
\end{equation}
\normalsize
In Eq. (\ref{eq:3}), $V^{0}$ represents the reference mesh with N vertices and $B^0 \in \mathbb{R}^{N \times M}$ is the B-spline coefficient matrix which expresses the influence of each control point on each vertex of the mesh. $P^0 \in \mathbb{R}^{M\times3}$ expresses the 3D coordinates of M control points. 
Once the coefficient matrix $B^0$ is computed using the reference mesh $V^0$, any mesh $V$ can be represented by deforming the control points as follows:

\begin{equation}\label{eq:5}
    V(\Delta P(I^i)) = B^{0}(P^{0}+\Delta P(I^i)),
\end{equation}
where $I^{i}$ indicates an arbitrary input image. Given a list of training data pairs of image $I^{i}$ and its corresponding ground truth mesh $V^{i}$, we train the neural network to extract image features and predict the appropriate deformation $\Delta P(I^i)$ of the control points from those features, so that the deformed mesh $V$ matches the ground truth mesh. All deformation of the mesh happens so that the vertex indices and triangle connectivity remain the same. The overview of our proposed method is shown in Fig. \ref{fig:2}.

\subsection{Applying Camera Projection to 3D Face Mesh}
Our model regresses pose parameters, as well as deformation parameters. It is a natural approach to consider the shape deformation in the world coordinate system and the camera projection separately. $T(s,R,t)$ indicates a 3D scaled orthographic projection ($3\times4$ affine transformation). The scale factor $s$ is applied to $x,y,z$ directions. This projection is called 3D in that it preserves the depths ($z$-values) of vertices, scaling them in the same factor as in the $x$ and $y$ directions. The transformation matrix $T(s,R,t)$ is multiplied to the deformed mesh to output the final transformed deformed mesh in the camera coordinate system.

\subsection{Loss Functions}

\subsubsection{Vertex Loss}
Since the ground truth mesh and the predicted mesh share the same topology, it is possible to calculate the distance between the vertices that share the same indices. The Mean Squared Error (MSE) of the vertex loss is defined as: 
\begin{equation}\label{eq:6}
   L_{Vertex} =\frac{1}{N}\sum_{i=0}^{N} (T(V(\Delta P(I^i))) - T_{g}(V^{i}))^2,
\end{equation}
where $N$ refers to the number of data, $V^{i}$ refers to the ground truth mesh of the training image $I^i$, and $V$ refers to Eq. (\ref{eq:5}) which outputs the deformed mesh from the predicted $\Delta P$. $T$ is a $3\times4$ transformation matrix and $T_{g}$ is the ground truth pose.

\subsubsection{Landmark Region Loss}

There exists a set of predefined vertex indices corresponding to 68 facial landmarks. If we simply use MSE, each facial region would be weighted depending on the number of landmarks in each region. In order to control the importance of each region, we divide the 68 landmarks into 9 different regions and form a weighted loss. The regions are divided into the left and right eyebrow, left and right eye, upper and lower nose, upper and lower lip, and contour. The loss of each landmark region, abbreviated as $LMRegion$, can be computed by sampling the mesh with landmark region indices, abbreviated as $LMIndex$ in the following equation.

\small
\begin{equation}\label{eq:7}
    L_{LMRegion} = \frac{1}{N} \sum_{i=0}^{N} \frac{1}{M} \sum_{j \in LMIndex} (T(V_{j}(\Delta P(I^i))) - T_{g}(V_{j}^{i}))^2,
\end{equation}
\normalsize
where $N$ refers to the number of data, $M$ refers to the number of landmarks of a specific region, and $V_j$ refers to the sampled vertices using $LMIndex$. The total loss can be computed as the weighted average of $L_{Vertex}$ defined in Eq. (\ref{eq:6}) and all $L_{LMRegion}$ defined in Eq. (\ref{eq:7}). In our implementation, the weight of the vertex loss is set to $0.46$, and the weight of each landmark region loss is set to $0.06$.

\section{EXPERIMENTS}

\begin{figure*}[t]
\centering
\includegraphics[width=0.65\linewidth]{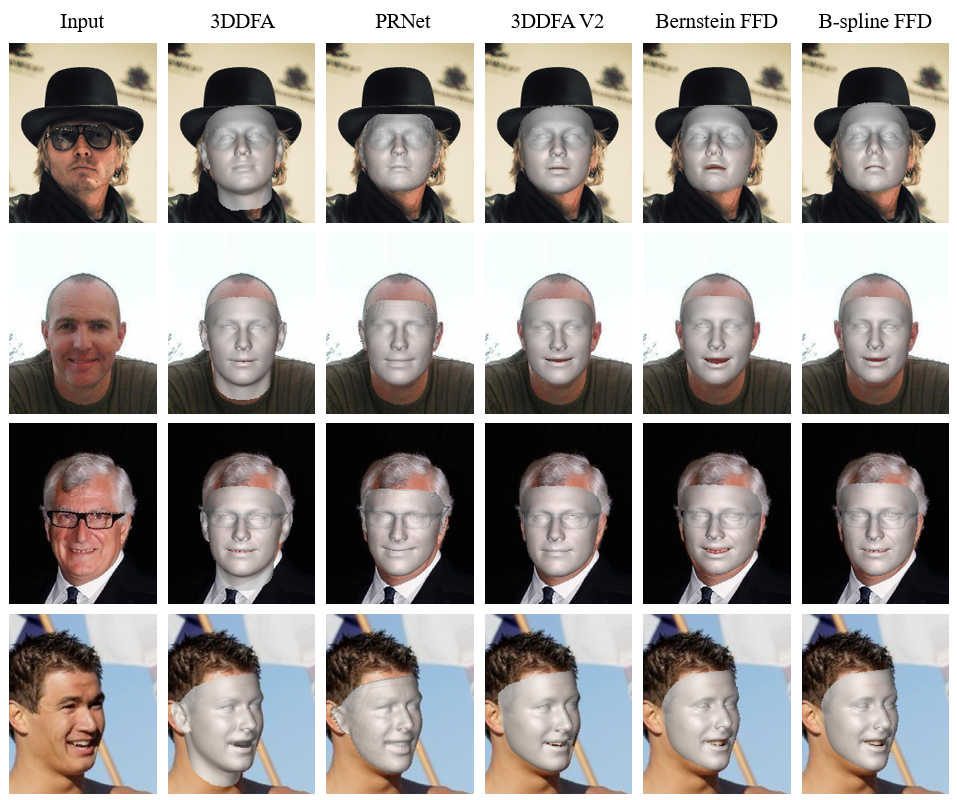}
\caption{Qualitative results of our method on CelebA compared to 3DDFA \cite{zhu2016face}, PRNet \cite{feng2018joint} and 3DDFA V2 \cite{guo2020towards}. Our results of two separate cases were reported, each using Bernstein FFD and B-spline FFD. B-spline FFD showed better results on the mouth area, since only considering neighbor control points and imposing local influence on vertices allow for finer control. B-spline FFD also outperformed other existing 3D face reconstruction methods.}\label{fig:3}
\end{figure*}

\subsection{Datasets and Protocols}
\subsubsection{300W-LP}
300W-LP \cite{zhu2016face} is composed of synthesized large-pose face images. The dataset consists of pairs of face images and 3DMM parameters, which were found by fitting a 3DMM built from a combination of the Basel Face Model \cite{paysan20093d} and FaceWarehouse \cite{cao2013facewarehouse}. 
We used an augmented version of 300W-LP \cite{3ddfa_cleardusk} for training, which has a more variety of extreme poses. Although the ground truth face meshes are based on 3DMM, our method has the capacity to generate deformed meshes beyond the linearly spanned space of 3DMM because the reconstructed mesh is not represented by 3DMM parameters, but by the deformation parameters.

\subsubsection{AFLW2000-3D}
AFLW2000-3D \cite{zhu2016face} contains the first 2,000 images of AFLW \cite{koestinger2011annotated}, with ground truth 3DMM parameters defined in a consistent manner with 300W-LP \cite{zhu2016face} and 68 3D landmarks. The dataset consists of images with large pose variations of yaw angles ranging from $-90\degree$ to $90\degree$ and with variations of illumination conditions and facial expressions. Thus, it is often used to evaluate the 3D face reconstruction performance on challenging in-the-wild images.

Quantitative evaluation of 3D face reconstruction for in-the-wild images is challenging due to the lack of pairs of 2D images and 3D models collected in an unconstrained setting. As an alternative, evaluation of facial landmarks can be indicative of the overall 3D face reconstruction accuracy, especially when considering both visible and invisible landmarks. 
We evaluated our method using facial landmark data of AFLW2000-3D. The protocol followed \cite{feng2018joint,guo2020towards,zhu2016face} to calculate the Normalized Mean Error (NME) normalized by the bounding box size. The result is reported by the range of yaw angles, which is divided into [0\degree, 30\degree], [30\degree, 60\degree] and [60\degree, 90\degree]. Following the works of \cite{feng2018joint,zhu2016face}, we randomly sampled 696 faces to balance the distribution so that each pose range has the same number of data.

\subsubsection{CelebA}
CelebA \cite{liu2015faceattributes} is a large-scale face dataset with over 200K in-the-wild images of celebrities. The images have large variations in pose and background clutter. We use the provided aligned images resized to $218 \times 178$ as our test data.

We use CelebA to conduct a qualitative evaluation of our model. As shown in Fig. \ref{fig:3}, reconstructed 3D face mesh is rendered on top of the input image using scaled orthographic projection. The rendered result shows the resemblance between the original face of the 2D image and the deformed face mesh.

\subsection{Implementation Details}
The training and experiments were implemented based on Pytorch \cite{NEURIPS2019_bdbca288}.
We employed a ResNet-50 architecture \cite{he2016deep} as the backbone of our network. At the end, we implemented a fully connected layer with 2,112 nodes to regress the deformation of 700 control points in 3D coordinates and 12 pose parameters for the transformation matrix. We divided the grid uniformly but with different dimensions on each axis $(l = 6, m = 19, n = 4)$.

\par
For training, we used an augmented version of the 300W-LP dataset \cite{3ddfa_cleardusk} with 687,854 images, which were obtained through applying random perturbations to pitch, yaw, and roll angles. Then, the images were cropped around the facial region and resized to $120\times120$. Finally, the images were normalized by the RGB mean and standard deviation.
We constructed the mesh data with 35,709 vertices covering the face area, excluding the ears and neck, using the provided 3DMM parameters. Our model was trained with the Adam optimizer using a learning rate of 0.001 and weight decay of 0.0005, for 50 epochs with a mini-batch size of 128.

\begin{table}[t]
\caption{Comparison between Bernstein FFD and B-spline FFD on AFLW2000-3D. The NME (\%) of facial landmarks with different yaw angles are reported.} 
\label{table:1}
\begin{center}
\begin{tabular}{|c|c|c|c|c|}
\hline
Method & 0\degree \ to 30\degree & 30\degree \ to 60\degree & 60\degree \ to 90\degree & Mean\\
\hline
Bernstein FFD & 2.97 & 3.70 & 4.90 & 3.86\\
\hline
B-spline FFD & 2.60 & 3.44 & 4.50 & 3.51\\
\hline
\end{tabular}
\end{center}
\end{table}

\subsection{Comparison between Bernstein FFD and B-spline FFD}
We have experimented with FFD using both Bernstein \cite{sederberg1986free} and B-spline basis functions \cite{hsu1992direct}, where the control points of Bernstein FFD have global influence on vertices while those of B-spline FFD have local influence. 
Empirically, we observed that one of the most challenging issues is properly controlling the mouth area, mostly due to the opened space between the lips. In fact, the qualitative results shown in Fig. \ref{fig:3} demonstrate that Bernstein FFD could hardly control the lips. 
We observed B-spline FFD yielded better results, since the human face is a deformable shape and requires local control. Furthermore, Table \ref{table:1} shows that B-spline FFD achieved higher accuracy on predicting facial landmarks in general. For fair comparison, both methods were tested with the same number of control points and grid dimensions.

\begin{table}[t]
\caption{Performance evaluation of 68 landmarks on AFLW2000-3D. The NME (\%) with different yaw angles are reported. The best results are highlighted.}
\label{table:2}
\begin{center}
\begin{tabular}{|c|c|c|c|c|}
\hline
Method & 0\degree \ to 30\degree & 30\degree \ to 60\degree & 60\degree \ to 90\degree & Mean\\
\hline
SDM \cite{xiong2015global} & 3.67 & 4.94 & 9.76 & 6.12\\
\hline
3DDFA \cite{zhu2016face} & 3.78 & 4.54 & 7.93 & 5.42\\
\hline
3DDFA + SDM \cite{zhu2016face} & 3.43 & 4.24 & 7.17 & 4.94\\
\hline
DeFA \cite{liu2017dense} & - & - & - & 4.50\\
\hline
Yu \emph{et al}. \cite{yu2017learning} & 3.62 & 6.06 & 9.56 & 6.41\\
\hline
3DSTN \cite{bhagavatula2017faster} & 3.15 & 4.33 & 5.98 & 4.49\\
\hline
3D-FAN \cite{bulat2017far} & 3.15 & 3.53 & 4.60 & 3.76\\
\hline
PRNet \cite{feng2018joint} & 2.75 & 3.51 & 4.61 & 3.62\\
\hline
CMD \cite{zhou2019dense} & - & - & - & 3.98\\
\hline
3DDFA V2 \cite{guo2020towards} & 2.63 & \textbf{3.42} & \textbf{4.48} & \textbf{3.51}\\
\hline
B-spline FFD & \textbf{2.60} & 3.44 & 4.50 & \textbf{3.51}\\
\hline
\end{tabular}
\end{center}
\end{table}

\subsection{Quantitative Results}
Table \ref{table:2} reports the NME of facial landmarks compared to existing methods including SDM \cite{xiong2015global}, 3DDFA \cite{zhu2016face}, DeFA \cite{liu2017dense}, Yu \emph{et al}. \cite{yu2017learning}, 3DSTN \cite{bhagavatula2017faster}, 3D-FAN \cite{bulat2017far}, PRNet \cite{feng2018joint}, CMD \cite{zhou2019dense}, and 3DDFA V2 \cite{guo2020towards}. Our method, B-spline FFD, showed better accuracy than most of the compared methods and achieved comparable performance to 3DDFA V2 \cite{guo2020towards}. It is worth noting that the landmarks are inherently a subset of 3DMM fitted meshes, so 3DMM-based methods tend to show better results.

\subsection{Qualitative Results}
Our reconstruction results of Bernstein FFD and B-spline FFD on the images of the CelebA dataset can be viewed at Fig. \ref{fig:3}, in comparison to 3DDFA \cite{zhu2016face}, PRNet \cite{feng2018joint}, and 3DDFA V2 \cite{guo2020towards}.
The qualitative evaluation demonstrates that our model is robust to large poses, not only frontal poses, as illustrated in the last row. In addition, it can reconstruct 3D meshes even when part of the face is occluded by glasses, as shown in the first and third row. Overall, this experiment validated that our model works well for in-the-wild images with various lighting, expressions, poses, and occlusions.

\section{CONCLUSION}
In this paper, we proposed a new approach to learning-based 3D face shape reconstruction through B-spline FFD. FFD can be considered as both model-based and model-free; the mesh is represented by low dimensional control points and polynomial basis functions, but still has no limit in expressiveness or model space. Moreover, the regressed deformation parameters are intuitive and self-explanatory. This further allows one to readily utilize the estimated mesh and control points for detailed adjustment and further deformation. Lastly, our model is meaningful in that it achieved comparable results to state-of-the-art methods of 3D face reconstruction from a single in-the-wild image.





\bibliographystyle{IEEEtran.bst}
\bibliography{root.bib}

\end{document}